%% file: root.tex
\documentclass[letterpaper, 10 pt, conference]{ieeeconf}  % Comment this line out if you need a4paper

\IEEEoverridecommandlockouts                              % This command is only needed if 
\usepackage{graphicx} % for pdf, bitmapped graphics files
\usepackage{amsmath} % assumes amsmath package installed
\usepackage[colorlinks,linkcolor=blue]{hyperref}
\hypersetup{
    colorlinks=true,
    linkcolor=blue,
    filecolor=blue,      
    urlcolor=blue,
    citecolor=blue,
}
\usepackage{makecell}
\usepackage{amsfonts}
\usepackage{balance}
\usepackage{url}
\usepackage{subfigure} 
\usepackage{booktabs}
\title{\LARGE \bf
Learning to Navigate in a VUCA Environment: \\Hierarchical Multi-expert Approach
}

\author{Wenqi Zhang$^{*,1}$, Kai Zhao$^{*,1}$ , Peng Li $^{\dag,1}$, Xiao Zhu$^{1}$, Faping Ye$^{1}$, Weijie Jiang$^{1}$, Huiqiao Fu$^{1}$ and Tao Wang$^{1,2}$ %<-this % stops a space

\thanks{*Denotes equal contribution: {\tt\small wenqizhang2018@gmail.com, kzhao@aiit.org.cn}} \thanks{$^{\dag}$Corresponding author: {\tt\small pli@aiit.org.cn}}

\thanks{$^{1}$Advanced Institute of Information Technology, Peking University.}
\thanks{$^{2}$Department of Computer Science and Technology, Peking University.}% stops a space

\thanks{This work was supported by State Key Laboratory of Computer Architecture(ICT, CAS) under Grant No. CARCHB202012.}}
\begin{document}
\maketitle
\thispagestyle{empty}
\pagestyle{empty}

\input{simple_abstract}
\input{simple_introduction}

\input{simple_related_work}

\input{simple_approach}

\input{simple_experiments}

\input{simple_conclusion}

%\input{acknowledgement}


%%%%%%%%%%%%%%%%%%%%%%%%%%%%%%%%%%%%%%%%%%%%%%%%%%%%%%%%%%%%%%%%%%%%%%%%%
%\input{appendix}
%\input{acknowledgement}

\bibliographystyle{IEEEtran}

\bibliography{IEEEabrv,IEEEexample}

\end{document}

%% file: simple_abstract.tex
\begin{abstract}

Despite decades of efforts, robot navigation in a real scenario with volatility, uncertainty, complexity, and ambiguity (VUCA for short), remains a challenging topic. Inspired by the central nervous system (CNS), we propose a hierarchical multi-expert learning framework for autonomous navigation in a VUCA environment. With a heuristic exploration mechanism considering target location, path cost, and safety level, the upper layer performs simultaneous map exploration and route-planning to avoid trapping in a blind alley, similar to the cerebrum in the CNS. Using a local adaptive model fusing multiple discrepant strategies, the lower layer pursuits a balance between collision-avoidance and go-straight strategies, acting as the cerebellum in the CNS. We conduct simulation and real-world experiments on multiple platforms, including legged and wheeled robots. Experimental results demonstrate our algorithm outperforms the existing methods in terms of task achievement, time efficiency, and security. A video of our results is available at \href{https://youtu.be/lAnW4QIWDoU}{https://youtu.be/lAnW4QIWDoU}.

\end{abstract}

%% file: simple_introduction.tex
\section{INTRODUCTION}

The robot navigation in a real environment has been extensively studied over decades, and significant progress has been made in recent works. Many indoor commercial service robots already have basic navigation capabilities. However, real-world emergencies usually require robots to perform detection or rescue tasks autonomously in a mapless place with dynamic obstacles and complex corridors. Navigation failures often occur under such emergent situations.

We investigated the key cause for navigation failures, and then divided environmental characteristics into four dimensions: Volatility(\emph{V}), Uncertainty(\emph{U}), Complexity(\emph{C}), Ambiguity(\emph{A}). The meaning of VUCA\footnotemark in navigation scenarios can be defined as follows:
\begin{itemize}
\item \emph{V}: highly dynamic objects may exist in a real scene. 
\item \emph{U}: environmental information is unknown beforehand, and the map can not be provided in advance.
\item \emph{C}: real environment may be complex, with lots of corridors and blind alleys.
\item \emph{A}: all robot's states must be acquired by the mounted sensors which are imperfect, not provided externally.
\end{itemize}

\begin{figure}[t]

	\centering
	\includegraphics[width = 1\linewidth]{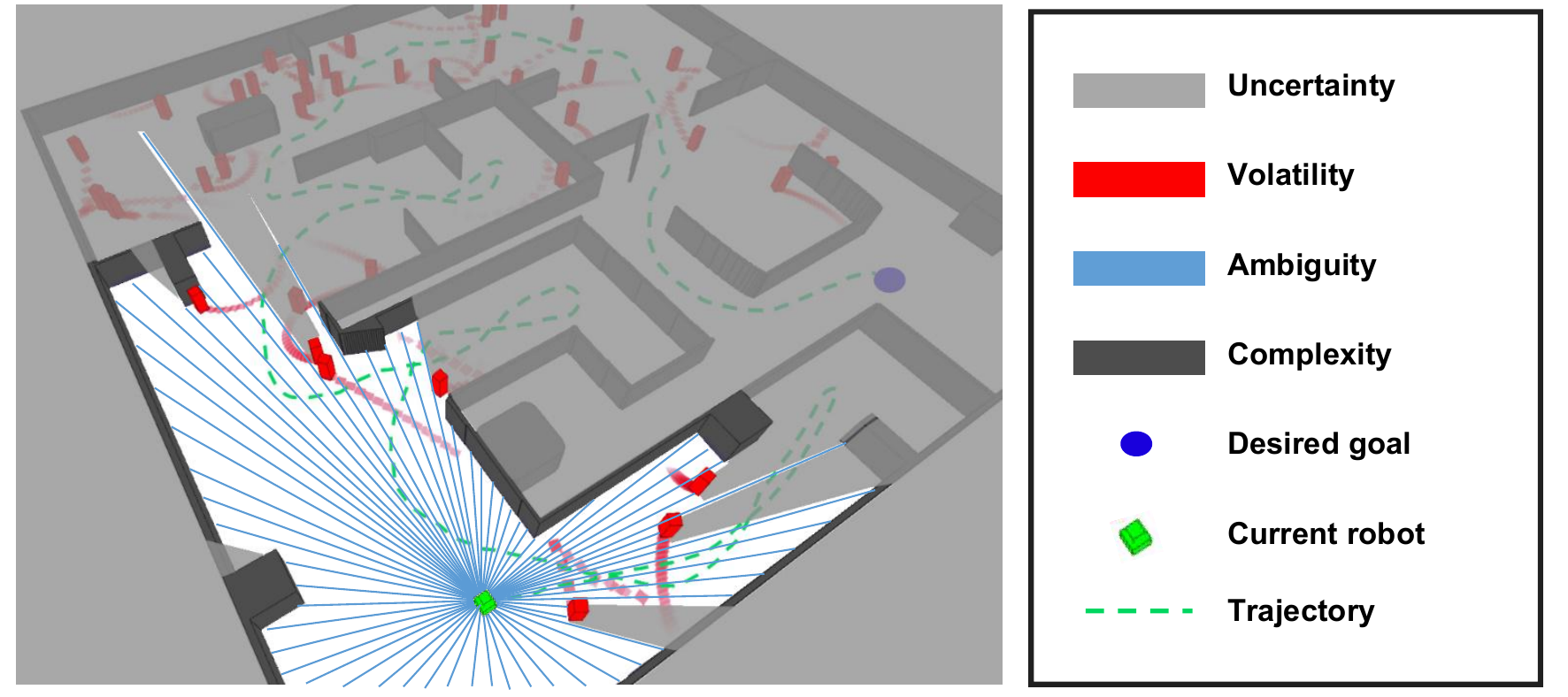}

\caption{In a completely strange and complicated scene without any map, it is a challenging task for the robot(green) to reach the goal(blue) and avoid moving obstacles(red). We propose a novel hierarchical navigation approach to handle such scenarios, including a value-based heuristic exploration mechanism and a local navigation approach by multi-expert fusion.}
	\label{fig:demo}
\end{figure}

As shown in Fig. \ref{fig:demo}, to perform the task in a typical VUCA environment successfully, robots need a reliable strategy, including efficient exploration, reasonable planning, and adaptive obstacle avoidance.

One of the commonly used navigation approaches is based on repeatedly mapping and re-planning trajectories\cite{dissanayake2001a,montemerlo2002fastslam}. These methods ensure to find a feasible path in the static scenario, but may not work well in the high-dynamic scenario. Specifically, moving obstacles may seriously hinder the effect of mapping and planning. Thus,  \emph{V} and \emph{A} challenges in VUCA greatly degrade the performance of these methods. As an alternative, learning-based methods, e.g., deep reinforcement learning (DRL), have been employed to address these problems in an end-to-end manner and greatly promoted the performance of obstacle avoidance in dynamic scenes\cite{chen2017decentralized,long2018towards}. However, these methods may lead robots to trap in a blind alley or a dead corner,  since they lack global map and environmental memory. Furthermore, the policy learned from interacting with the environment is similar to an instinctive response to a simple situation, short of the abilities to cope with complicated environments. 
As a result, they may not solve \emph{C} and \emph{U} problems in a VUCA environment independently. Thus, those existing methods mentioned above cannot tackle the navigation problem in a VUCA environment well.

\footnotetext{VUCA originated from the military field since the 1990s.}

To the best of our knowledge, the autonomous navigation problem in a fully VUCA environment has not been completely resolved in previous work. Inspired by the design of the biological central nervous system(CNS), different brain regions are responsible for different levels of motion control. The cerebrum is in charge of the overall planning of movement, meanwhile, the cerebellum is focusing on movement coordination and balance. Motivated by this, we propose a hierarchical navigation framework. Similar to CNS, the upper layer model, acting as the cerebrum, has the abilities of long-term memory and reasoning, while the lower layer model is comparable to the cerebellum with instantaneous memory and rapid strain instinct. That is to say, the upper layer guides robots to move toward the targets, and the lower layer produces adaptive behaviors in response to changing situations. Considering VUCA, the upper layer can handle the problems of \emph{C} and \emph{U}, by realizing mapping, exploration, and path planning. And the lower layer enhances the ability of obstacle avoidance, to alleviate the challenges caused by \emph{V} and \emph{A}. The experimental results on a real quadruped robot and a wheeled robot demonstrate that our approach is more reliable and flexible to achieve navigation tasks for an autonomous robot in a VUCA environment.

To sum up, the contributions of this paper are as follows:
\begin{itemize}
\item We propose a hierarchical framework in a real VUCA environment, which jointly resolves global exploration under complex unknown scenes and local adaptive navigation in a dynamic environment, promoting generalization ability in practice.

\item A novel multi-expert learning algorithm is proposed to enhance navigation ability, generating adaptive behaviors by fusing pre-trained strategies to alleviate the challenges caused by \emph{V} and \emph{A} in a VUCA environment.

\item Both in simulation and real-world, autonomous navigation of legged and wheeled robots are achieved, showing the effectiveness of our method in a VUCA environment.

\end{itemize}

%% file: simple_related_work.tex
\section{RELATED WORK}
\subsection{Map-based and Optimization-based methods}
Conventional methods, e.g. navigation toolkit in ROS (Robotics Operate System)\cite{movebase}, usually divide the processes of mapping, localization,  planning, and obstacle avoidance into different modules. The robot needs to search for global paths as well as local paths to avoid dynamic obstacles\cite{dissanayake2001a}. SLAM-based approach, including Laser-based\cite{dissanayake2001a,montemerlo2002fastslam} and Vision-based\cite{davison2007monoslam,jung2003high} algorithms, is mainly used for mapping and localization, and the planning algorithm is adopted to tackle the question of planning and obstacle avoidance. With the development of computer vision, employing visual information to navigate has also made remarkable progress\cite{guerrero2005visual,chen2019a}. However, due to the laser scan that can be used in most scenarios, the Laser-based approaches are more suitable for navigation in a VUCA environment.

In addition to traditional search algorithms, e.g., Dijkstra and A*, there are other types of planning algorithm, including artificial potential field method\cite{khatib1986real}, genetic algorithm\cite{carbone2008an}, ant colony algorithm\cite{dorigo2004ant}, and multi-robots optimization-based approaches \cite{yu2016optimal,preiss2017downwash,augugliaro2012generation}. These methods formulate motion planning as a search or optimization problem. The optimization target is usually to find the shortest path to reach the goal without collision. However, these planning algorithms need to model the environment in a limited space. That is to say, the methods mentioned above must rely on a given map and cannot independently solve VUCA navigation problems without external information provided, such as the speed and position of all robots. 

\subsection{Learning-based method}
Recently, deep reinforcement learning (DRL)\cite{mnih2015human} has been employed to settle navigation problem in an end-to-end manner and achieved considerable successes\cite{pfeiffer2018reinforced,wu2018learn,wu2020bnd}. Tai et al.\cite{tai2016towards} trained a DQN agent for obstacle avoidance in a simulated indoor environment. They also extended this approach to a mobile robot with continuous control\cite{tai2017virtual}. Many noteworthy works using DRL for collision avoidance in multi-robot\cite{long2018towards,chen2017decentralized} and dense crowd environment\cite{fan2018crowdmove}  were published. A hierarchical structure\cite{wang2018learning} was reported to solve navigation problems in a complex environment. Most of these methods owned similar feedback by designing an elaborate reward mechanism. Additionally, work\cite{fan2018fully} adopted a hybrid strategy to trade off efficiency and safety, including PID control, DRL strategy, and safe strategy. The work\cite{semnani2020multi} divided the environment into four situations, and corresponding strategies could be switched automatically. 

In the field of social-aware navigation, recent works have used DRL to learn socially cooperative policies\cite{chen2017socially,everett2018motion}. These methods take into account the motion state of obstacles and enhance the navigation ability. SARL\cite{chen2019crowd} demonstrated a superior ability to generate crowd-aware navigation policies. Another improved approach\cite{li2019sarl}, called SARL*, introduced a dynamic local goal-setting mechanism and a map-based safe action space. However, these learning-based methods have some limits in practice, lacking capabilities of long-term memory and global planning, and they are easy to overfit a specific training scenario. That is to say, they can not solve \emph{C} and \emph{U} problems in a VUCA environment.
%(可以补充更多的works)

\begin{figure}[t]

	\centering
	\includegraphics[scale=0.5,width = 1\linewidth]{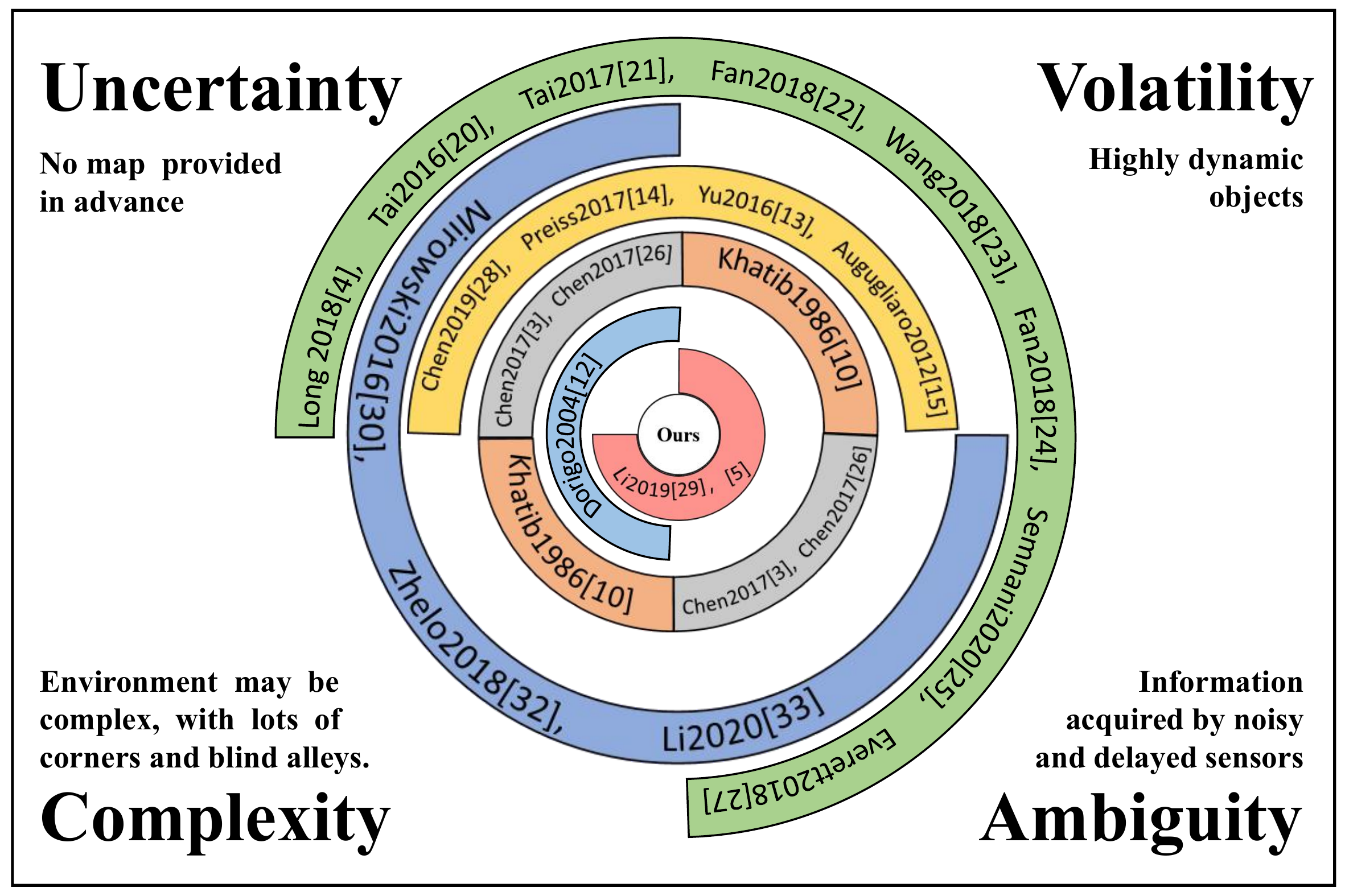}

	\caption{A simple classification of typical works using VUCA as a metric. Previous works have solved partial challenges in the VUCA environment.}
	\label{fig:related}
\end{figure}

Another challenge that needs to be considered in reality is how to explore an unfamiliar environment efficiently. Mirowski\cite{mirowski2016learning} introduced an approach using long short-term memory to assist the agent to acquire historical memory and implicitly understand the surrounding. Pathak\cite{pathak2017curiosity} proposed an intrinsic curiosity module(ICM), and it drove the agent to fit the environmental model, which was used to calculate curiosity reward. Then Oleksii et al.\cite{zhelo2018curiosity} applied this approach in navigation, prompting the robot to explore the unseen scene and avoiding falling into a blind alley. Li\cite{li2020deep} proposed an exploration algorithm based on DDQN which took the partial slam-map as input to improve exploration efficiency and adaptability in an unknown environment.
%(可以补充更多的works)

Fig. \ref{fig:related} shows the classification of several typical works using VUCA as a metric.  Significant progress has been made for navigation in the partial VUCA environment. However, the high dynamic environment and imperfect sensors greatly degrade the performance of the map-based and optimization-based methods, and also the learning-based method lacks global planning ability. To our best knowledge, previously methods are not fully applicable to navigation tasks in a VUCA environments without delicate tuning and retraining.

%% file: simple_approach.tex
\section{APPROACH}

\subsection{A hierarchical navigation framework} 
About VUCA, the challenges of \emph{V} and \emph{A} bring collision risk to robots, and the key to solving \emph{U} and \emph{C} is long-term memory and efficient global planning.

\subsubsection{Exploration and navigation mechanism}
\begin{figure}[t]

	\centering
	\includegraphics[scale=0.4,width = 1\linewidth]{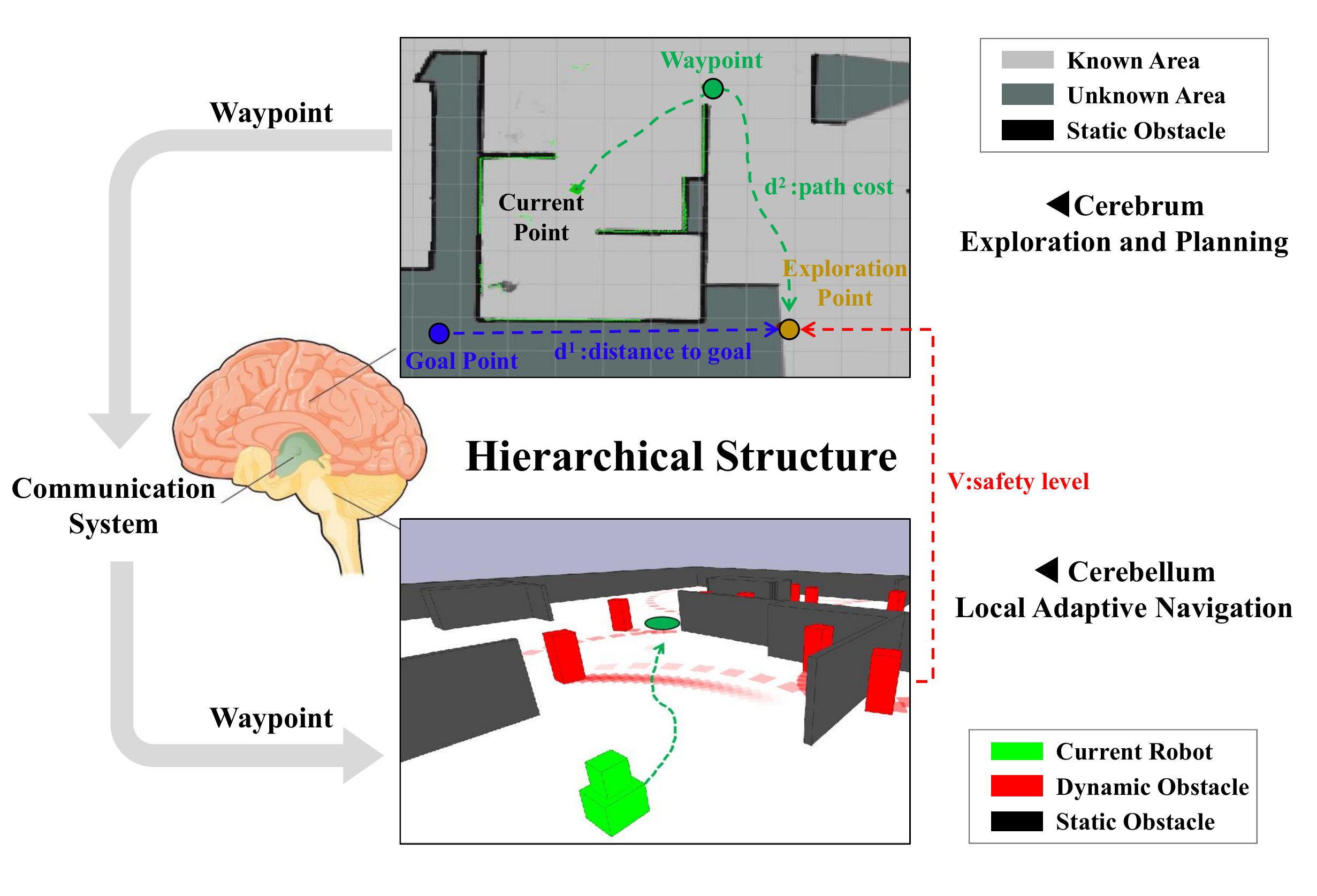}
 
	\caption{The goal point(blue) lies in the unknown area, and the robot is trapped in the blind alley. Considering the path cost, distance to goal, and safety level, the cerebrum(the upper layer module) calculates the exploration point (orange) and the waypoint(green) to drive robots out of the corner, and the cerebellum(the lower layer module) is responsible for short-distance navigation toward waypoint and moving obstacles avoidance.}

	\label{fig:brain}
\end{figure}

We design a hierarchical structure to address these challenges. The hierarchical design is shown in  Fig. \ref{fig:brain}. As shown, it contains two layers, the cerebrum in the upper layer is used to implement simultaneous exploration of unknown areas and path planning. The cerebellum in the lower layer is used for short-distance navigation. We define the current robot position as $P_{current}$, the goal position as $P_{goal}$, and $P_{exploration}$ for next step exploration. Considering two types of heuristic factors, the cerebrum selects the most reasonable $P_{exploration}$ and then plan a set of safety waypoints $P_{waypoint}$ from  $P_{current}$ to $P_{exploration}$. The cerebellum treats $P_{waypoint}$ from the cerebrum as a temporary goal, and adopts an adaptive strategy to move towards this goal. The upper is to solve the problems of \emph{U} and \emph{C}, and the lower aims to alleviate the challenges of \emph{V} and \emph{A}.

\subsubsection{Framework} 
\begin{figure*}[t]
	\centering
	\includegraphics[width = 1\linewidth]{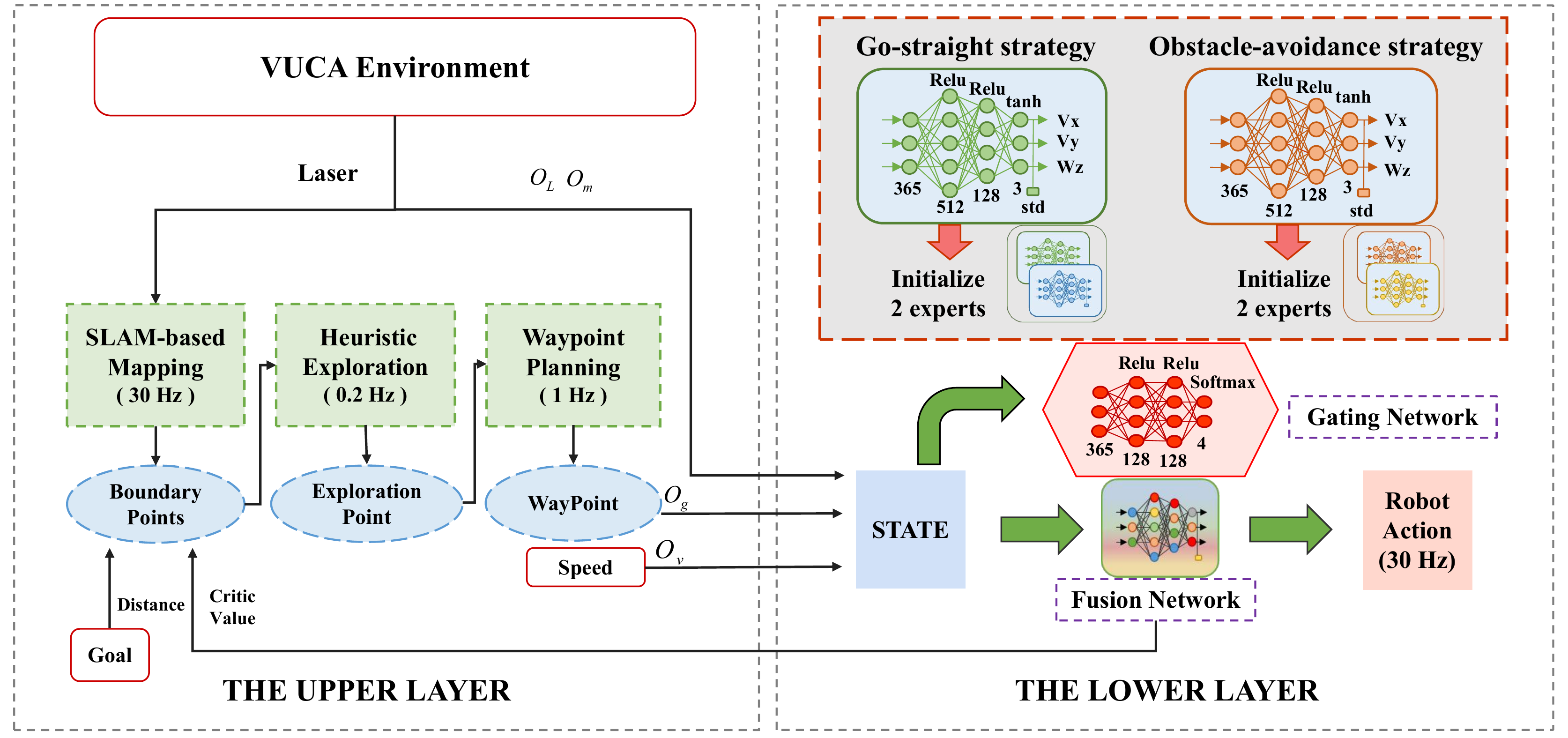}
	\caption{The framework of our method, containing a value-based heuristic exploration mechanism at the upper layer and a multi-expert fusion approach at the lower layer. In the upper layer, SLAM-based mapping, heuristic exploration, and waypoint planning modules are designed to search for feasible path points in the unknown environment towards goal. In the lower layer, we propose a multi-expert learning approach that contains multiple navigation experts, each with a unique navigation ability, e.g., some experts know how to behave towards goal quickly in static scenarios, and the others are skilled in avoiding dynamic obstacles. Then we adopt a gating network to fuse multiple expert networks dynamically into a versatile and adaptive model.}
	%   \Description{}
	\label{fig:pipeline}
\end{figure*}

The overall framework is given in Fig. \ref{fig:pipeline}. In the lower layer, inspired by \cite{yang2020multi}, a multi-expert learning approach contains multiple experts, each with a unique navigation ability, and a gating network is applied to fuse multiple experts dynamically into a more versatile and adaptive policy. The heuristic exploration mechanism in the upper layer includes SLAM-based mapping, heuristic exploration, and waypoint planning modules. The three modules are responsible for updating the existing map, selecting the best exploration region toward the destination, and calculating reasonable waypoints, respectively. Each component runs at a certain frequency. The mapping, exploration, and planning modules run at 30\,Hz, 0.2\,Hz, and 1\,Hz, respectively. Considering highly dynamic circumstances, the frequency of the adaptive model in the lower layer is 30\,Hz.
 
\subsection{Cerebellum-local adaptive navigation}

To better tackle \emph{V} and \emph{A} challenges, we analyze the strategies required for navigation, which can be divided into two categories: \emph{go-straight} strategy and \emph{obstacle-avoidance} strategy. If only \emph{obstacle-avoidance} strategy is adopted, robots incline to make a detour around obstacles, and sometimes even temporarily move away from the goal, resulting in a longer path and cost-time. By contrast, the \emph{go-straight} navigation strategy drives robots to approach the goal constantly and robots may try to pass through the obstacles gap, leading to a dangerous situation. Therefore, robots must learn to trade off these two strategies to cope with changing scenarios. It is difficult for an end-to-end DRL agent to learn both strategies at the same time, as agent may forget the strategies they learned before. Thus, we design a multi-expert fusion method through a two-stage training process.

\subsubsection{Deep reinforcement learning components} 
We use the PPO algorithm\cite{schulman2017proximal} for training and design compact state space, practical action space, and reasonable reward functions. The observation vector of the robot can be denoted as: $[O_L^t,O_m^t,O_g^t,O_v^t]$. Specifically, the robot can observe surroundings $O_L$ from laser scans and relative location $O_g$ to the goal in the robot’s frame. In order to capture the changes of the environment, we also calculate relative motion from historical laser observation {$O_L^t$, \ldots , $O_L^{t-n}$}, denoted as $O_m=\sum_{k=1}^n(O_L^t-O_L^{t-k})/k$. $O_m$ integrates the motion of the agents themselves and external obstacles. Besides, the agent's velocity is also provided, namely $O_v$. We design action vector as follows: $[v_x,v_y,w_z]$, which represent forward velocity, lateral velocity, and yaw angular velocity respectively. As shown in Table \ref{table:rewardform1}, seven reward functions: ${R_g,R_o,R_c,R_r,R_t,R_e,R_a}$ are introduced for training. Each expert requires a specific subset of reward components. The two processes are training in ROS Stage\cite{stage_ros}, which is a lightweight robot simulator.

\begin{table}[htbp]
\caption{Reward terms table}
\begin{center}
\begin{tabular}{ccccc}
\toprule
 
{Terms} & {Formula} & \makecell[c]{Go-\\straight} & \makecell[c]{Obstacle-\\avoidance} & \makecell[c]{Fusion \\network}   \\ \midrule
\begin{tabular}[c]{@{}c@{}}\end{tabular}        {$R_g$} & {$w_g\cdot(d^t-d^{t-1})$}& {$w_g$=3} & {$w_g$=1} & {$w_g$=4} \\ \midrule
{$R_o$} & $\frac{w_o\cdot\max(0.6-\min O_L^t,0)}{0.6-\max(0.6-\min O_L^t,0)}$ & $w_o$=0 & $w_o$=-0.4 & $w_o$=0\\ \midrule
{$R_c$} & $w_c\cdot15$ when collision & $w_c$=-0.25 & $w_c$=-1 & $w_c$=-1\\ \midrule
{$R_r$} & $w_r\cdot20$ when reach goal & $w_r$=1 & $w_r$=0.25 & $w_r$=1\\\midrule
{$R_t$} & $w_t\cdot0.01$ & $w_t$=-1 & $w_t$=1 & $w_t$=0 \\\midrule
{$R_e$} & $w_e\cdot5$ when  timeout & $w_e$=-1 & $w_e$=1 & $w_e$=0\\\midrule 
{$R_a$} & $w_a\cdot \max(\omega_z-0.3,0)$ & $w_a$=-0.5 & $w_a$=0 & $w_a$=0\\ \bottomrule
\end{tabular}
\end{center}
\label{table:rewardform1}  
\end{table}

\subsubsection{First training stage-two discrepant experts} 
\begin{figure}[htbp]
\centering

\subfigure[]{
\begin{minipage}[t]{0.4\linewidth}
\flushleft
\includegraphics[height=2.8cm,width=2.8cm]{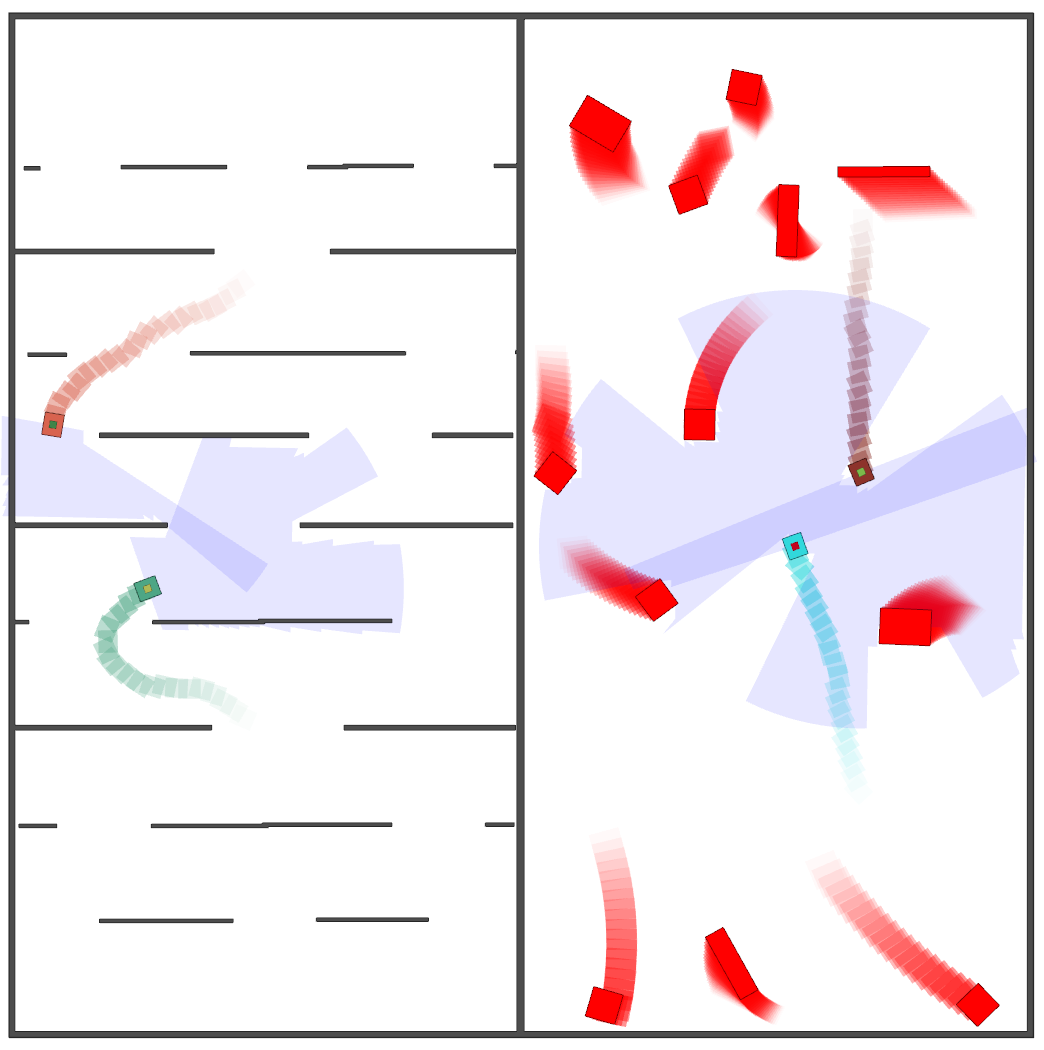}
%\caption{}
\end{minipage}%
}%
\subfigure[]{
\begin{minipage}[t]{0.52\linewidth}
\centering
\includegraphics[width=1\linewidth]{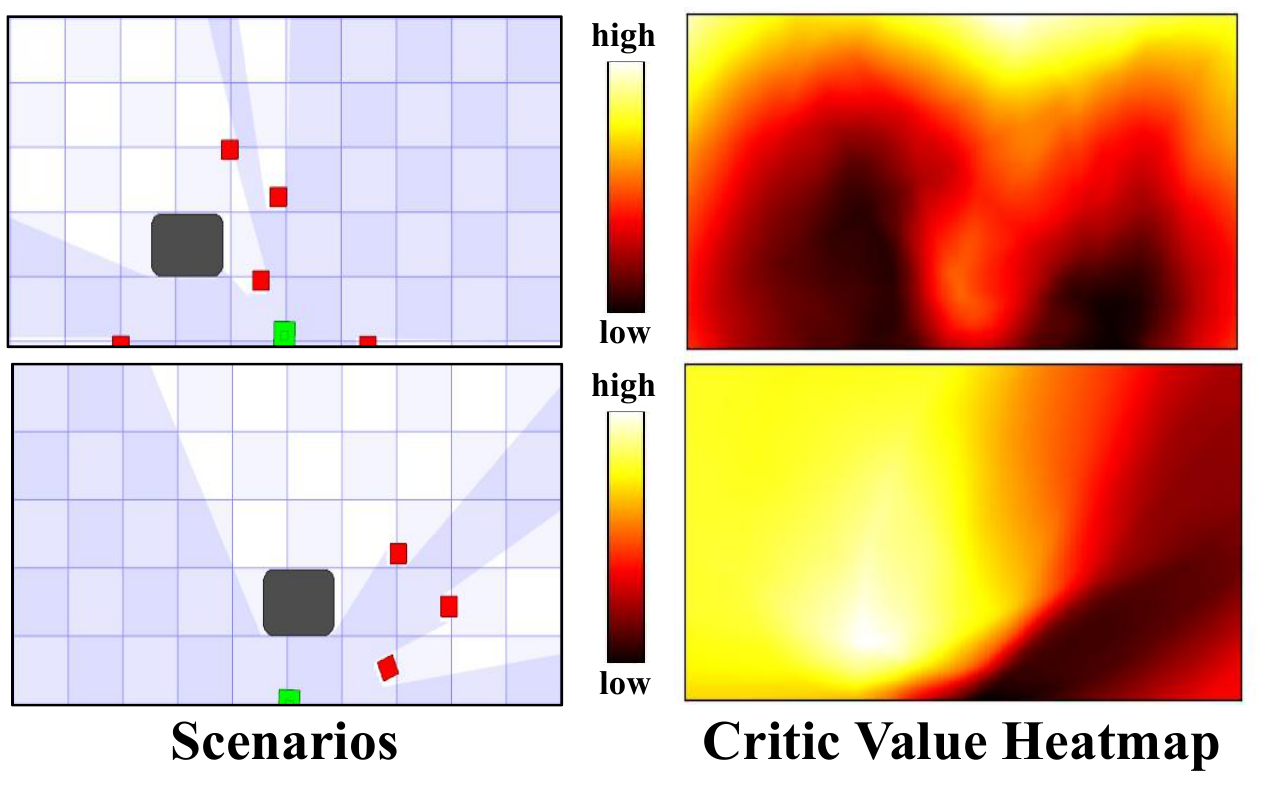}
%\caption{fig2}
\end{minipage}%
}%
 
\centering
\caption{(a) two training scenarios. The left is for the \emph{go-straight} strategy, while the right is used to train the \emph{obstacle-avoidance} strategy. (b) we use heatmaps to visualize the safety-levels around robots. The safety-level is calculated by the critic network in the multi-expert fusion model. The critic network takes all points around the robot as goals input ${O_{g}}$, and calculates the critic output of each point. Results 
indicate open areas and obstacle-free spaces have higher safety-level. }
\label{fig:scenario}
\end{figure}

In the first training stage, we train two expert strategies, one for \emph{obstacle-avoidance} in a dynamic environment, and the other for \emph{go-straight} navigation in a static environment. Fig. \ref{fig:pipeline} depicts the structure of expert networks. The state input, action output, and network structure of the two expert models are completely identical, and the only two distinctions are training scenarios and reward functions. Fig. \ref{fig:scenario}(a) shows two types of scenarios. The left is the static scenario, including massive walls and roadblocks, and the right is the dynamic scenario, including plenty of obstacles with different shapes, which walk randomly. Additionally, two reward function settings can be seen in Table \ref{table:rewardform1}. The \emph{obstacle-avoidance} strategy pays more attention to the threat of obstacles, so we add $R_o$ to its reward function, while \emph{go-straight} navigation strategy abhors excessive angular velocity, so we add $R_a$.

\subsubsection{Second training stage-fusing multiple experts} 

In the second training stage, the gating network learns to fuse all experts. Considering that navigation in complex scenes requires at least four skills: \emph{forward}, \emph{turn}, \emph{emergency stop}, and \emph{backward}. We initialize four networks by copying parameters from the two trained experts model before, i.e., two expert networks are replicated into four networks(corresponding to four skills), and then a gating network fuses all parameters of four networks. The output of the gating network is non-linearly mapped by \emph{softmax} function to obtain four weighting coefficients. We co-train four networks and gating network in different tasks. Fig. \ref{fig:pipeline} also depicts the structure of the fusion network and the training procedure. In detail, we use distributed-DRL(DPPO) to train model in multiple environments at the same time. The reward function of this stage is also shown in the last column of Table \ref{table:rewardform1}.   

Similar to \cite{yang2020multi}, let $x$,$y$,$h$ denote the dimensions of the input, output, and hidden layer, respectively. Let $W$ be the network's weights. The parameter of the fusion network is
\begin{equation}
\begin{split}
\Psi_{f}= \{ W_0 \in \mathbb{R}^{x \times h1 },W_1 \in \mathbb{R}^{h1 \times h2 },W_2 \in \mathbb{R}^{h2 \times y } \}
\end{split}
\end{equation}
and the parameter set of each expert network is  
\begin{equation}
\begin{split}
\Psi_{e}^n= \{ W_0^n \in \mathbb{R}^{x \times h1 },W_1^n \in \mathbb{R}^{h1 \times h2},W_2^n \in \mathbb{R}^{h2 \times y } \}
\end{split}
\end{equation}the weights $W$ are fused as 
\begin{equation}
\begin{split}
W_i=\sum_{n=1}^4\alpha_n W_i^n
\end{split}
\end{equation}
where $n=1,\ldots,4$ is the index of experts, $i=0,1,2$ is the index of the network layer, and $\alpha_n \in [0,1] $ are weighting coefficients generated by the gating network, and the same operation is applied to bias $B$. Through co-training, four trained experts have been evolved, and gating network has also learned to output different weights to integrate all experts according to different situations.

%Both strategies are aware of obstacles, but the \emph{obstacle-avoidance} strategy is conservative than \emph{go-straight} strategy.
%The heatmaps indicate open areas and obstacle-free spaces have higher safety-level.
 
\subsection{Cerebrum-exploration and planning}
In the absence of an initial map, robots need to explore and navigate simultaneously. When some blind alleys exist in the environment, robots may be away from the goal temporarily to find a way out. However,  lacking historical memory, robots may repeatedly wander in a limited location. In this section, as shown in Fig. \ref{fig:pipeline}, we adopt a value-based heuristic exploration policy to generate waypoints through three modules: SLAM-based mapping, heuristic exploration, and waypoint planning. The purpose of ours value-based heuristic policy is to find an exploration point in frontier. This frontier exploration method ensures that a feasible path from the starting to the goal can be found in a limited map.  

\subsubsection{SLAM-based mapping} 
We employ the widely used laser SLAM algorithm Cartographer\cite{hess2016real} for localization, mapping, and boundary-searching. The resolution of the occupancy grid map is adjusted according to the size of the robot, and each pixel in the map has a range of 0 to 1, which represents the probability of being occupied by obstacles. We classify the pixel $x_i$ in the map $M(x)$ into three categories: occupied($M(x_i)>0.52$), free($M(x_i)<0.48$), and unknown($M(x_i)\in[0.48, 0.52]$). All free points adjacent to unknown points, are defined as $P_{boundary}$.

\subsubsection{heuristic exploration} 
In order to select the best exploration point $P_{exploration}$ from candidate points $P_{boundary}$.
We introduce two heuristic factors to design the heuristic function. As shown in Fig. \ref{fig:brain}, the heuristic function is
\begin{equation}
\begin{split}
i^*= \mathop{\arg\min}_{i} (\frac{D^h_i-D^h_{min}}{D^h_{max}-D^h_{min}}+\gamma \frac{V^h_{max}-V^h_i}{V^h_{max}-V^h_{min}} ) \label{exploration}
\end{split}
\end{equation}
The first term is the goal heuristic factor($D^h$) based on goal distance and path cost, and the second term is the safety-level heuristic factor($V^h$). The coefficient $\gamma$ is to trade off these two terms, and the two denominators are used to normalize both items. Where $D^h_i=d^1_i+d^2_i$, $d^1_i$ represents the Euclidean distance between the $i$\,th $P_{boundary}$ and $P_{goal}$, and $d^2_i$ refers to the length of path from $P_{boundary}^i$ to $P_{current}$. $V^h_i$ is the output of the critic network in the multi-expert fusion model(Section III B) when setting the $P_{boundary}^i$ as $O_g$. Inspired by work\cite{fan2019getting}, we use the critic network of the fusion model to evaluate the safety level of all candidate points. As shown in Fig. \ref{fig:scenario}(b), we produce heatmaps to show the safety-level of the entire space surrounding the robots. Considering the distance to the goal, the path cost, and the safety-level of the dynamic environment, two heuristic factors(i.e. $D^h$ and $V^h$) are combined to select the best exploration point.

Besides, $P_{exploration}$ is re-selected immediately when the following situations occur:
\begin{itemize}
\item The map changes a lot during exploration process. We use the cross-entropy to measure the change of the map. The entropy of the map is
\begin{equation}
H(m)=-\mathop{\sum}_{i}\mathop{\sum}_{j}p(m_{i,j})\log p(m_{i,j})
\end{equation}
$p(m_{i,j})$ is the occupied probability of the cross grid of $i$-th row and $j$-th column.
\item The robot reaches the $P_{exploration}$.
\item The state of $P_{exploration}$ has changed.
\item The state of $P_{goal}$ changes from unknown to known.
\end{itemize}

\subsubsection{waypoint planning}
After selecting a suitable $P_{exploration}$, this module plans a feasible path from $P_{current}$ to $P_{exploration}$. To reduce the communication burden between the upper and lower layers, only some sparse waypoints $P_{waypoint}$ are selected from the planned path. Being set as the goal point $O_g$ in the fusion model, $P_{waypoint}$ is the farthest point to $P_{current}$ and no static obstacles are lying on the straight line between them. As shown in  Fig. \ref{fig:brain}, robots would quickly leave the blind alley to avoid repeatedly wandering in the same position.

%% file: simple_experiments.tex
\section{EXPERIMENTS AND EVALUATION}
In this section, we quantitatively compare the proposed hierarchical approach with other methods in various simulated VUCA scenarios. We also demonstrate the performance improvement of fusion strategy compared to two expert strategies. Real-world experiments using different robot platforms are conducted in a real VUCA environment.

\begin{figure*}[t]
	\centering
	\includegraphics[width = 1\linewidth]{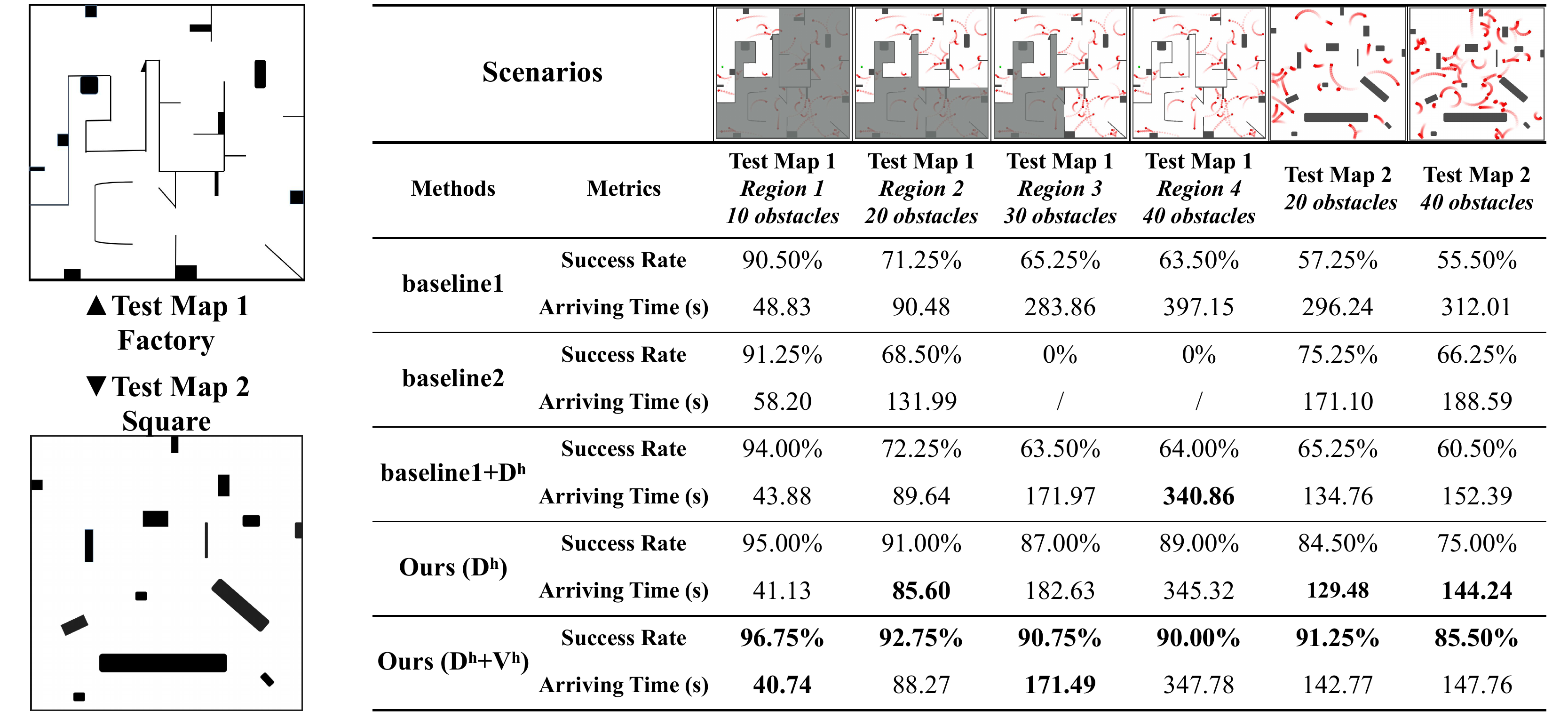}
	\caption{Evaluation results of our system. The first factory scenario includes 40 dynamic obstacles and complicated blind alleys. We divide the factory scenario into four sub-regions. Region 1 is the simplest and Region 4 is the most complex for navigation. Region 1 to 4 represents a quarter, two quarters, three quarters, and the entire area of the factory, respectively. The second square scenario contains plenty of high-speed moving obstacles and roadblocks.
}
	%   \Description{}
	\label{fig:experiment}
	
\end{figure*}
\textbf{Baselines}: we select two methods for comparison, denoting \emph{move\_base}\cite{movebase} as baseline1, learning-based method\cite{long2018towards} as baseline2. We refer to our full model as Ours($D^{h}+V^{h}$), which means it contains two heuristic factors(Section III C) for exploration, and the model without safety-level heuristic factor is denoted as Ours(${D^{h}}$), which means it only contains ${D^{h}}$ as the heuristic factor. Normally, baseline1 requires a given map to run. For a fair comparison, we use an initial unknown map to test baseline1 and then update the map to re-plan the path simultaneously. And goal heuristic factor ${D^{h}}$ is added to baseline1, denoted as baseline1${+D^{h}}$. Note that, since the original baseline2 algorithm is not open-sourced, we use the reproduced implementation\footnote{https://github.com/Acmece/rl-collision-avoidance} as a substitute.

\textbf{Metrics}: We design five metrics for evaluation:
\begin{itemize}
\item \emph{Success rate}(\%) 
\item \emph{Arriving time} (when the robot reaches the goal)(sec)
\item \emph{Crash rate}(\%) 
\item \emph{ARSPS} measures the average risk scores per step. The calculation of risk score for each step ($S_{risk}^{t}$) is similar to the reward function $R_o$(Section III B).

\item \emph{ANSPS} evaluates the average navigation scores per step.  

%\emph{ANSPS} and \emph{ARSPS} denote as 
\begin{equation}
{S_{risk}^{t}}=\frac{\max(0.6-\min O_L^t,0)}{0.6-\max(0.6-\min O_L^t,0)}  
\end{equation}
\begin{eqnarray}
ARSPS=\frac{\sum{S_{risk}^{t}}}{N_{steps}},ANSPS=\frac{D_{start} - D_{end}}{D_{start} \cdot N_{steps}} 
\end{eqnarray}

${N_{steps}}$ means the total steps for this episode, ${D_{start}}$ and ${D_{end}}$ represent the Euclidean distances between the robot and the goal at the beginning and the end.
\end{itemize}
\subsection{Contrast experiments of the system}
\textbf{Test scenarios}: To investigate the performance of our approach, we design two types of simulated scenarios (i.e. factory and square). The first factory scene is mainly used to test exploration and navigation capabilities, and the second square scene is applied to test obstacle avoidance capabilities in highly dynamic environments. The maximum speed of the obstacle is twice that of a robot in the square scene.

We test two baselines with multiple parameters and record the best result of each baseline. For each method, every test case is evaluated 400 times. The test scenarios and corresponding results are given in Fig. \ref{fig:experiment}. Results show our method can safely reach the goal in a relatively short time and has a higher success rate for all scenarios, while baseline1 and baseline2 are prone to be trapped and fail in complex scenes(e.g. Region 3 and 4) and often collide in high dynamic scenes (e.g. Test Map 2). Specifically, in the simple scenario (e.g. Region 1), all five methods perform well, but in the complex scenario (e.g. Region 3 and 4), baseline2 cannot reach the goal, since this method lacks global planning ability. Baseline1 usually collides in high dynamic scenes (e.g. Test Map 2). Compared with baseline1, both baseline1${+D^{h}}$ and Ours(${D^{h}}$) methods can significantly reduce arriving time and improve exploration efficiency. Moreover, Ours(${D^{h}}$) has a higher success rate than baseline1${+D^{h}}$, showing its advantages in task achievement.

\begin{figure}[t]

	\centering
	\includegraphics[scale=0.4,width = 1\linewidth]{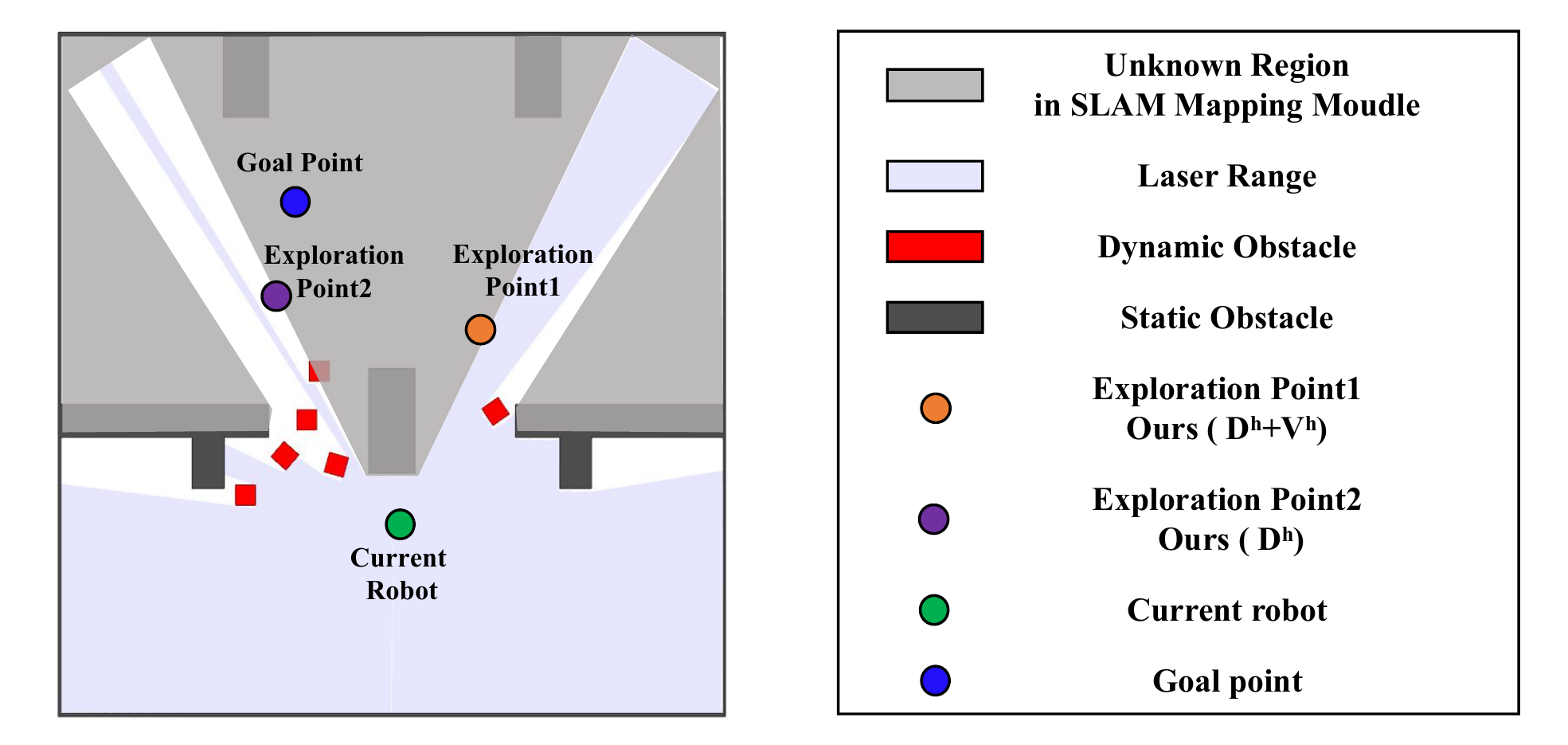}
	\caption{We test the robot in double-branch-structure scene. The left branch has more dynamic obstacles than the right. Although the goal point(blue) is close to the left branch, the upper layer would choose the right exploration point(orange) instead of the left one(purple) in most cases. Due to the safety-level factor($V^{h}$), the probability of collisions is greatly reduced.}
	\label{fig:exploration_point_test}
\end{figure}

Besides, in comparison to Ours($D^{h}$), Ours($D^{h}+V^{h}$) obtains higher success rate in all cases. While sacrificing time efficiency in some scenes, Ours(${D^{h}}+{V^{h}}$) has a significant improvement in task achievement. It is proved that critic value as a heuristic factor can effectively improve the security of the exploration process. As shown in Fig. \ref{fig:exploration_point_test}, to better research the effect of safety-level heuristic factor($V^{h}$), we test the robot in a double-branch-structure environment with different numbers of obstacles. The results show robots tend to choose the side with fewer obstacles as exploration points.

\subsection{Ablation experiments for multi-expert fusion approach}
\begin{figure}[t]

	\centering
	\includegraphics[scale=0.4,width = 1\linewidth]{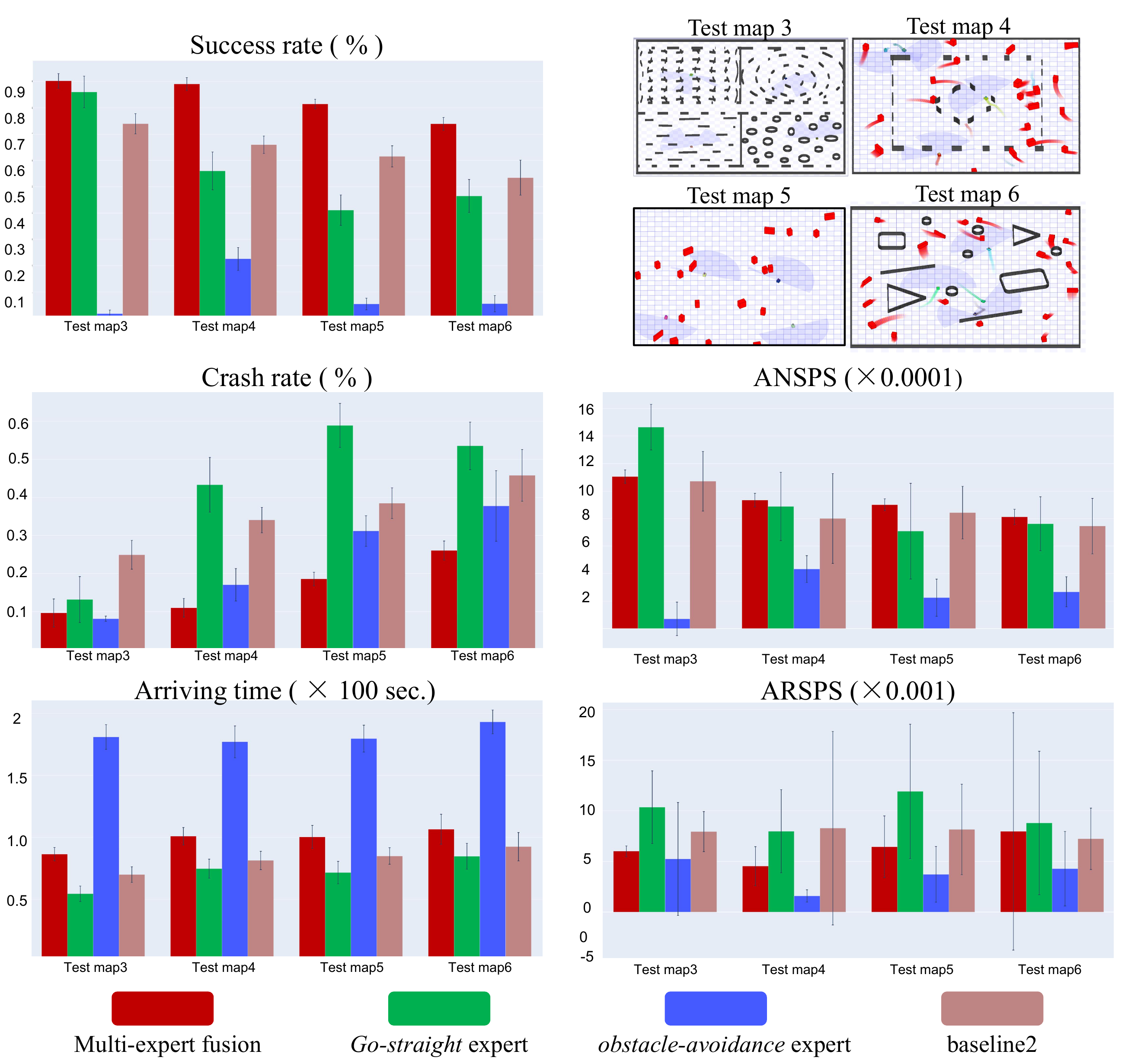}

	\caption{Comparison among four approaches in four test scenarios. The results show our fusion approach outperforms baseline2 and two expert methods in terms of success rate or arriving time. The multi-expert fusion model combines the advantages of two trained expert strategies.}

	\label{fig:zhuzhuangtu}
\end{figure}

To evaluate the effectiveness of the multi-expert fusion approach, we present the performance of the fusion model and compare it with two expert strategies and baseline2. Although the fusion model has a slightly longer arriving time than baseline2(Fig. \ref{fig:zhuzhuangtu}), it obtains a higher success rate and a lower crash rate. Fig. \ref{fig:zhuzhuangtu} also displays the characteristics of two expert strategies, the \emph{ANSPS} of \emph{go-straight} strategy is higher, with a shorter arriving time but a higher crash rate. In contrast, the \emph{ARSPS} of \emph{obstacle-avoidance} strategy is better, with a lower crash rate but a lower success rate. The multi-expert fusion model performs well in all metrics. It is noteworthy that both two trained expert strategies are imperfect at first, with obvious shortcomings, i.e. \emph{go-straight} strategy pursues navigation efficiency, but its obstacle avoidance ability is insufficient, and \emph{obstacle-avoidance} strategy has a strong collision avoidance ability, but the time efficiency of that is relatively low. However, the multi-expert fusion model has achieved higher navigation efficiency and stronger obstacle avoidance capabilities after co-training. It indicates that our approach learns the advantages of two trained experts and enables robots to cope with different situations flexibly.

\begin{figure}[t]

	\centering
	\includegraphics[scale=0.4,width = 1\linewidth]{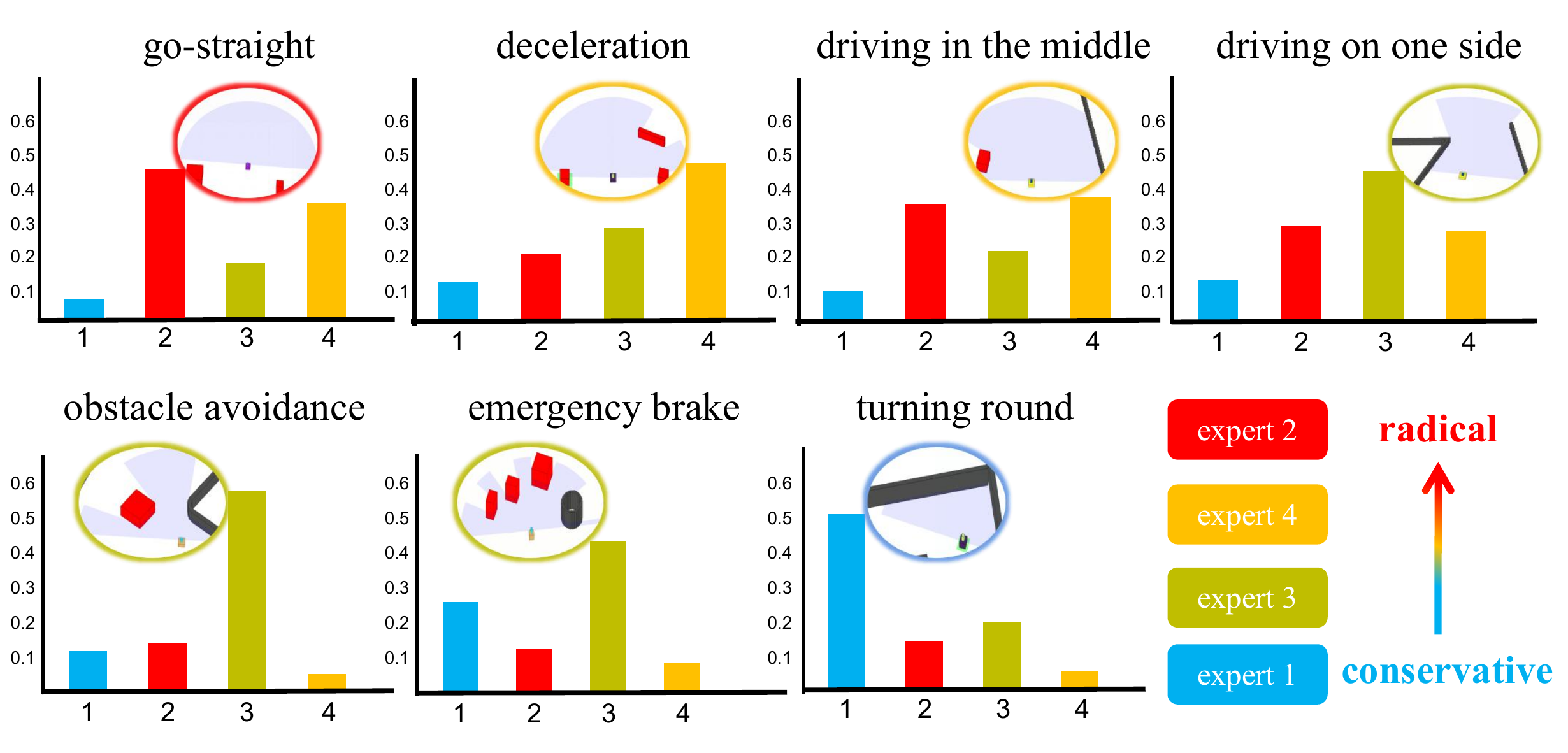}

	\caption{The output of the gating network changes according to different observations. Circular areas show various scenes during navigation. The gating network selects different experts to implement specific behaviors. The behavior of expert 1 is the most conservative, while expert 2 is the most radical. The gating network has learned to generate variable weights for all experts in response to the observation.}
	\label{fig:gate}
\end{figure}

We also analyze the output of the gating network. Fig. \ref{fig:gate} shows the correlations between the outputs of the gating network and different navigation behaviors, and it reveals each behavior has a dominant expert strategy. After co-training, the gating network selects the corresponding expert strategies according to the different observations, e.g. expert 2 and 4 are selected for navigation straightly in an obstacle-free scene, while expert 1 tends to bypass obstacles cautiously, and expert 3 prefers emergency brake and rapid collision avoidance, as shown in Fig. \ref{fig:gate}. This phenomenon verifies that our design can fuse various strategies.

\subsection{Real-world experiments in different robot platforms}
\begin{figure}[t]

	\centering
	\includegraphics[scale=0.4,width = 1\linewidth]{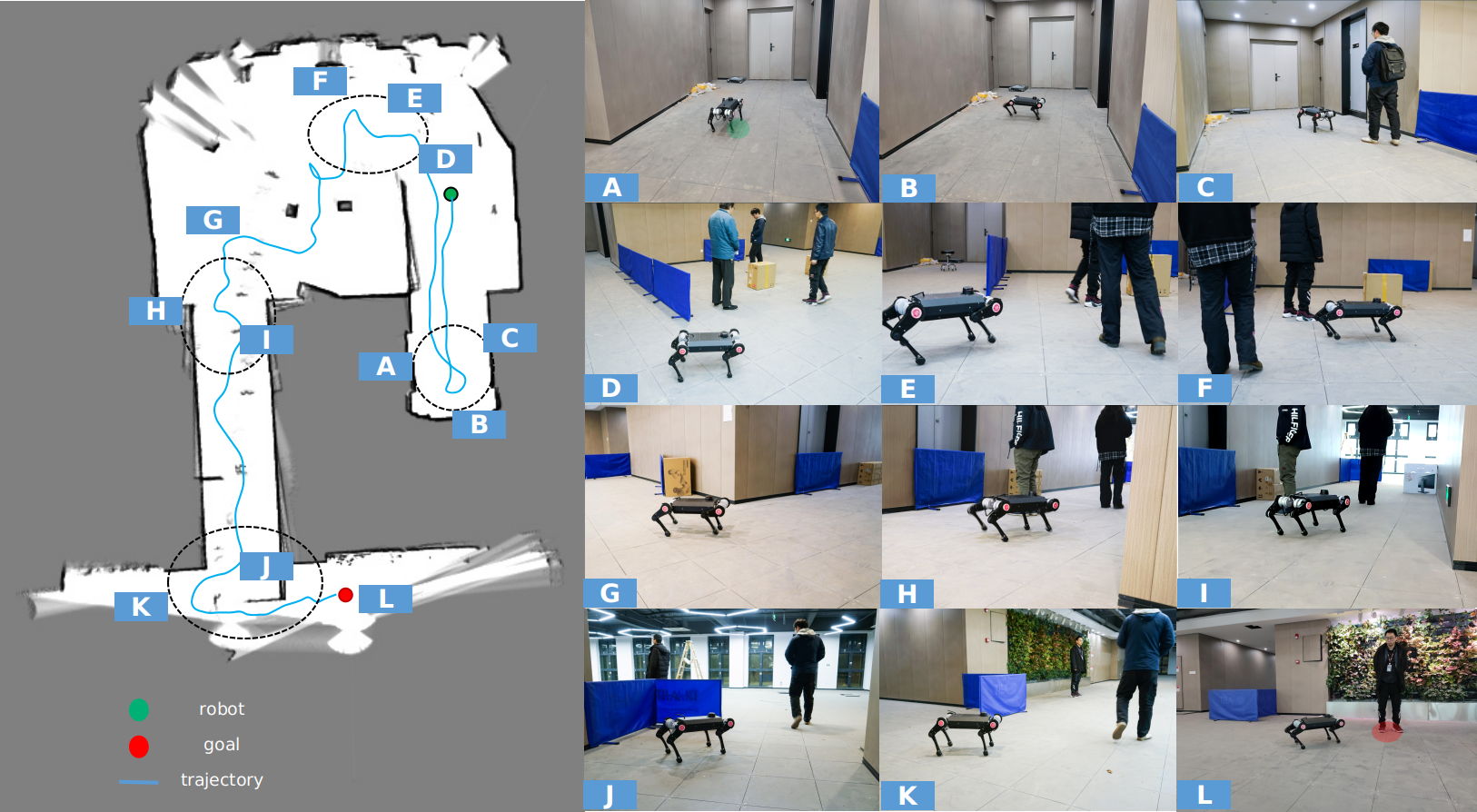}

	\caption{A real-world experiment using a quadruped robot.}

	\label{fig:realdog}
\end{figure}

To verify our approach in more realistic scenarios, we conduct real-world experiments, using a quadruped and a wheeled robot in a complicated building with a dense crowd. Similar to the rescue site, test scenario is a real VUCA environment, with many mobile pedestrians and complex architectural layouts. Fig. \ref{fig:realdog} exhibits our robot platforms and navigation trajectory. Experiments display our approach has excellent exploration and navigation abilities in all cases.

%% file: simple_conclusion.tex
\section{CONCLUSIONS}
In this paper, we develop a hierarchical multi-expert learning framework for autonomous navigation in a VUCA environment. Considering complex layouts and potential obstacles in an unknown environment, our framework includes two different functional models in the upper and lower layers. The upper layer is responsible for global exploration and planning, while the lower layer is in charge of local navigation and obstacle avoidance. Specifically, the upper module adopts a heuristic exploration mechanism to guide robots to search for a goal in an unfamiliar environment, and a multi-expert fusion approach in the lower layer is utilized to guide robots to safely reach the goal at the same time. Both simulation and real-world experiments show our method outperforms the existing methods in terms of task achievement, time efficiency, and security.